\documentclass[10pt,twocolumn,letterpaper]{article}

\usepackage{cvpr}              

\usepackage{graphicx}
\usepackage{amsmath}
\usepackage{amssymb}
\usepackage{booktabs}

\usepackage{array}
\usepackage{cite}

\usepackage{mdwmath}

\usepackage{pifont}

\usepackage{bbm}
\usepackage{dsfont}
\usepackage{bm}
\usepackage{comment}
\usepackage{enumerate}  
\usepackage{mathrsfs}
\usepackage{makeidx}         
\usepackage{epsfig,latexsym}
\usepackage{psfrag}
\usepackage[ruled,vlined]{algorithm2e}
\SetKwInput{KwIn}{Require}

\usepackage{color}
\usepackage{colortbl}
\definecolor{lightgray}{rgb}{.93,.93,.93}

\newcommand{\mf}{\mathbf}

\newcommand{\mr}{\mathrm}

\usepackage[pagebackref,breaklinks,colorlinks,linkcolor=blue]{hyperref}

\usepackage[capitalize]{cleveref}
\crefname{section}{Sec.}{Secs.}
\Crefname{section}{Section}{Sections}
\Crefname{table}{Table}{Tables}
\crefname{table}{Tab.}{Tabs.}

\def\cvprPaperID{2956} 
\def\confName{CVPR}
\def\confYear{2022}

\newenvironment{myitemize}{\begin{list}{$\bullet$}
{\setlength{\topsep}{1mm}
\setlength{\itemsep}{0.25mm}
\setlength{\parsep}{0.25mm}
\setlength{\itemindent}{0mm}
\setlength{\partopsep}{0mm}
\setlength{\labelwidth}{15mm}
\setlength{\leftmargin}{4mm}}}{\end{list}}

\begin{document}

\title{Federated Class-Incremental Learning}

\author{Jiahua Dong\textsuperscript{1, 2}\footnotemark[1]~, Lixu Wang\textsuperscript{3}\footnotemark[1]~, Zhen Fang\textsuperscript{4}, Gan Sun\textsuperscript{1}\footnotemark[2]~, Shichao Xu\textsuperscript{3}, Xiao Wang\textsuperscript{3}\footnotemark[2]~, Qi Zhu\textsuperscript{3}\footnotemark[2]\\
\textsuperscript{1}State Key Laboratory of Robotics, Shenyang Institute of Automation, Chinese Academy of Sciences.\\
\textsuperscript{2}University of Chinese Academy of Sciences. \textsuperscript{3}Northwestern University.\\
\textsuperscript{4}DeSI Lab, AAII, University of Technology Sydney.\\
{\tt\small dongjiahua@sia.cn,\;lixuwang2025@u.northwestern.edu,\;\{fzjlyt,\;sungan1412\}@gmail.com}\\
{\tt\small \{shichaoxu2023@u.,\;wangxiao@,\;qzhu@\}northwestern.edu}}


%

\maketitle
\renewcommand{\thefootnote}{\fnsymbol{footnote}}
\footnotetext[1]{Equal contributions (ordered alphabetically). \footnotemark[2]Corresponding authors.}

\begin{abstract}
Federated learning (FL) has attracted growing attentions via data-private collaborative training on decentralized clients. However, most existing methods unrealistically assume object classes of the overall framework are fixed over time. It makes the global model suffer from significant catastrophic forgetting on old classes in real-world scenarios, where local clients often collect new classes continuously and have very limited storage memory to store old classes. Moreover, new clients with unseen new classes may participate in the FL training, further aggravating the catastrophic forgetting of global model. To address these challenges, we develop a novel \underline{G}lobal-\underline{L}ocal \underline{F}orgetting \underline{C}ompensation (GLFC) model, to learn a global class-incremental model for alleviating the catastrophic forgetting from both local and global perspectives. 
Specifically, to address local forgetting caused by class imbalance at the local clients, we design a class-aware gradient compensation loss and a class-semantic relation distillation loss to balance the forgetting of old classes and distill consistent inter-class relations across tasks. To tackle the global forgetting brought by the non-i.i.d class imbalance across clients, we propose a proxy server that selects the best old global model to assist the local relation distillation. Moreover, a prototype gradient-based communication mechanism is developed to protect the privacy.
Our model outperforms state-of-the-art methods by 4.4\%$\sim$15.1\% in terms of average accuracy on representative benchmark datasets. 
The code is available at \url{https://github.com/conditionWang/FCIL}.

\end{abstract}

\section{Introduction}\label{sec:intro}
Federated learning (FL)~\cite{DBLP:journals/corr/abs-1910-07796, wang2021addressing, 8945292, pmlr-v119-karimireddy20a} enables multiple local clients to collaboratively learn a global model while providing secure privacy protection for local clients. It successfully addresses the data island challenge without completely compromising clients' privacy~\cite{Lange_2020_CVPR, NEURIPS2020_24389bfe}. Recently, it has attracted significant interests in academia and achieved remarkable successes in various industrial applications, \emph{e.g.}, autonomous driving~\cite{8917592}, wearable devices~\cite{DBLP:journals/corr/abs-1804-07474}, medical diagnosis~\cite{10.1145/3394486.3403070, What_Transferred_Dong_CVPR2020} and mobile phones~\cite{DBLP:journals/corr/abs-1906-04329}.

Generally, most existing FL methods~\cite{wang2021addressing, 10.1145/3394486.3403070, DBLP:journals/corr/abs-1910-07796, hong2021federated} are modeled in a static application scenario, where data classes of the overall FL framework are fixed and known in advance.  However, real-world applications are often dynamic, where local clients receive the data of new classes in an online manner. To handle such a setting, existing FL methods typically require storing all training data of old classes at the local clients' side so that a global model can be obtained via FL, however the high storage and computation overhead may render the FL unrealistic when new classes arrive dynamically~\cite{DBLP:journals/corr/abs-1906-04329, 10.1145/3394486.3403070, DBLP:journals/corr/abs-1804-07474, wang2022nontransferable}. And if these methods~\cite{10.1145/3394486.3403070, DBLP:journals/corr/abs-1910-07796} are required to learn new classes continuously with very limited storage memory, they may suffer from significant performance degradation (\emph{i.e.}, catastrophic forgetting~\cite{Kirkpatrick3521, 10.5555/3294996.3295059, Rebuffi_2017_CVPR}) on old classes. Moreover, in real-world scenarios, new local clients that collect the data of new classes in a streaming manner may want to participate in the FL training, which could further exacerbate the catastrophic forgetting on old classes in the global model training. 


To address these practical scenarios, we consider a challenging FL problem named \textit{\underline{F}ederated \underline{C}lass-\underline{I}ncremental \underline{L}earning (FCIL)} in this work. In the FCIL setting, each local client collects the training data continuously with its own preference, while new clients with unseen new classes could join in the FL training at any time. More specifically, the data distributions of the collected classes across the current and newly-added clients are non-independent and identically distributed (non-i.i.d.). FCIL requires these local clients to collaboratively train a global model to learn new classes continuously, with constraints on privacy preservation and limited memory storage~\cite{Rebuffi_2017_CVPR, wu2019large}. To better comprehend the FCIL problem, we here use COVID-19 diagnosis among different hospitals as a possible example~\cite{dayan2021federated}. Imagine that before the pandemic, there could be hundreds of hospitals working collaboratively to train a global infectious disease diagnosis model via FL. Due to the sudden emergence of COVID-19, these hospitals will collect a large amount of new data related to COVID-19 and add them into the FL training as new classes. Moreover, new hospitals whose main focus is not infectious diseases may join the fight against COVID-19, where they have little data of the old infectious diseases, and all hospitals should learn to diagnose the old diseases and new COVID-19 variants. In such scenarios, most existing FL methods will likely suffer from catastrophic forgetting on old diseases diagnosis under the sudden emergence of new COVID-19 variants data.

An intuitive way to address new classes (\emph{e.g.}, learning new COVID-19 variants) continuously in the FCIL setting is to simply integrate FL~\cite{8945292, DBLP:journals/corr/abs-1910-07796, DBLP:journals/corr/McMahanMRA16} and class-incremental learning (CIL)~\cite{Rebuffi_2017_CVPR, Yan_2021_CVPR, Hu_2021_CVPR} together. However, such strategy needs the central server to know when and where the data of new classes arrives (privacy-sensitive information), which violates the requirement of privacy preservation in FL. In addition, although local clients can utilize conventional CIL~\cite{Rebuffi_2017_CVPR, Hu_2021_CVPR} to address their local catastrophic forgetting, the non-i.i.d. class imbalance across clients may still cause heterogeneous forgetting on different clients, and this simple integration strategy could further exacerbate local catastrophic forgetting due to the heterogeneous global catastrophic forgetting on old classes across clients.

To tackle these challenges in FCIL, we propose a novel \textit{\underline{G}lobal-\underline{L}ocal \underline{F}orgetting \underline{C}ompensation (GLFC)} model in this paper, which effectively addresses local catastrophic forgetting occurred on local clients and global catastrophic forgetting across clients. Specifically, we design a class-aware gradient compensation loss to alleviate the local forgetting brought by the class imbalance at the local clients via balancing the forgetting of different old classes, and propose a class-semantic relation distillation loss to distill consistent inter-class relations across different incremental tasks. To overcome the global catastrophic forgetting caused by the non-i.i.d. class imbalance across clients, we design a proxy server to select the best old global model for the class-semantic relation distillation at the local side. Considering the privacy preservation, the proxy server collects perturbed prototype samples of new classes from local clients via a prototype gradient-based communication mechanism, and then utilizes them to monitor the performance of the global model for selecting the best one. Our model achieves 4.4\%$\sim$15.1\% improvement in terms of average accuracy on several benchmark datasets, when compared with a variety of baseline methods. The major contributions of this paper are summarized as follows:

\begin{myitemize}
 \item We address a practical FL problem, namely Federated Class-Incremental Learning (FCIL), in which the main challenges are to alleviate the catastrophic forgetting on old classes brought by the class imbalance at the local clients and the non-i.i.d class imbalance across clients.
 \item We develop a novel Global-Local Forgetting Compensation (GLFC) model to tackle the FCIL problem, alleviating both local and global catastrophic forgetting. 
 To our best knowledge, this is the first attempt to learn a global class-incremental model in the FL settings. 
 \item We design a class-aware gradient compensation loss and a class-semantic relation distillation loss to address local forgetting, by balancing the forgetting of old classes and capturing consistent inter-class relations across tasks.
 \item We design a proxy server to select the best old model for class-semantic relation distillation on the local clients to compensate global forgetting, and we protect the communication between this proxy server and clients with a prototype gradient-based mechanism for privacy. 
\end{myitemize}

\section{Related Work}
\textbf{Federated Learning} (FL) is a decentralized learning framework that can train a global model by aggregating local model parameters~\cite{pmlr-v97-yurochkin19a, DBLP:conf/iclr/WangYSPK20,DBLP:conf/iclr/YangFL21,DBLP:conf/iclr/LiJZKD21}. To collaboratively learn a global model, \cite{DBLP:journals/corr/McMahanMRA16} proposes to aggregate local models via a weight-based mechanism. \cite{DBLP:journals/corr/abs-1812-06127} introduces a proximal term to help local model approximate the global ones. \cite{DBLP:journals/corr/abs-1910-07796} focuses on minimizing the model discrepancies across clients via an improved EWC. Moreover, \cite{8945292} designs a layer-wise aggregation strategy to reduce computation overhead \cite{DBLP:journals/pami/ZhangHWXP21, DBLP:journals/pami/ZhangCGMXLP21}. \cite{pmlr-v119-karimireddy20a} sacrifices the local optimality for rapid convergence, while \cite{NEURIPS2020_24389bfe, Lange_2020_CVPR} aim to improve the performance of local models. \cite{DBLP:conf/iclr/PengHZS20} integrates unsupervised domain adaptation \cite{DMCDAAA, Liu_2021_ICCV, Liu_2022_CVPR, 9616392_Dong, DBLP:conf/ijcai/ZhangLFY0020} into federated learning framework \cite{DBLP:conf/icml/HamerMS20, lyu2021novel}.
However, these existing FL methods cannot effectively learn new classes continuously, due to the limited memory to store old classes at the local clients' side.

\textbf{Class-Incremental Learning} (CIL) aims to learn new classes continuously while tackling forgetting on old classes \cite{Ahn_2021_ICCV,DBLP:conf/aaai/KimC21, zhang2020knowledge}. Without access to the data of old classes, \cite{Kirkpatrick3521} designs new regulators for balancing the biased model optimization caused by new classes, and~\cite{10.1007/978-3-319-46493-0_37, Shmelkov_2017_ICCV} use the knowledge distillation to surmount catastrophic forgetting. \cite{10.5555/3294996.3295059, NIPS2018_7836} introduce generative adversarial networks to produce synthetic data of old classes. As claimed in \cite{Rebuffi_2017_CVPR, wu2019large, liu2020mnemonics, 10.1007/978-3-030-58565-5_6}, the class imbalance between old and new classes is a crucial challenge for exemplar replay methods. \cite{Yan_2021_CVPR, Liu2020AANets} design a self-adaptive network to balance biased predictions. \cite{Hu_2021_CVPR} uses the causal effect on knowledge distillation to rectify class imbalance. \cite{Christian2021MGeoCont} introduces geodesic path to traditional knowledge distillation. \cite{Ahn_2021_ICCV} combines task-wise knowledge distillation and separated softmax for bias compensation. These CIL methods however cannot be applied to tackle our FCIL problem, due to their strong assumptions on when and where the data of new classes arrive.

\section{Problem Definition}
In the standard class-incremental learning \cite{Rebuffi_2017_CVPR, Christian2021MGeoCont, Shmelkov_2017_ICCV}, there are a sequence of streaming tasks $\mathcal{T} = \{\mathcal{T}^t\}_{t=1}^T$, where $T$ denotes the task number, and the $t$-th task $\mathcal{T}^t = \{\mf{x}_i^t, \mf{y}_i^t\}_{i=1}^{N^t}$ consists of $N^t$ pairs of samples $\mf{x}_i^t$ and their one-hot encoding labels $\mf{y}_i^t\in \mathcal{Y}^t$. $\mathcal{Y}^t$ represents the label space of the $t$-th task including $C^t$ new classes that are different from $C^p=\sum_{i=1}^{t-1}C^i \subset \cup_{j=1}^{t-1}\mathcal{Y}^j$ old classes in previous $t-1$ tasks. Inspired by \cite{Rebuffi_2017_CVPR, wu2019large, liu2020mnemonics}, we construct an exemplar memory $\mathcal{M}$ to select $\frac{|\mathcal{M}|}{C^p}$ exemplars for each old class in the $t$-th incremental task, and it satisfies $\frac{|\mathcal{M}|}{C^p} \ll \frac{N^t}{C^t}$.

We then extend conventional class-incremental learning to Federated Class-Incremental Learning (FCIL). Given $K$ local clients $\{\mathcal{S}_l\}_{l=1}^K$ and a global central server $\mathcal{S}_G$, for the $r$-th  global round ($r=1,\cdots, R$), a set of local clients are randomly selected to participate in the gradient aggregation. Specifically, once the $l$-th client $\mathcal{S}_l$ is selected at each global round for the $t$-th incremental task, it will receive the latest global model $\Theta^{r, t}$, and train $\Theta^{r, t}$ on its privately accessible $t$-th incremental task $\mathcal{T}_l^t\cup\mathcal{M}_l\sim\mathcal{P}_l^{|\mathcal{T}_l^t|+|\mathcal{M}_l|}$, where $\mathcal{T}_l^t=\{\mf{x}_{li}^t, \mf{y}_{li}^t\}_{i=1}^{N_l^t}\subset \mathcal{T}^t$ is the training data of new classes, $\mathcal{M}_l$ denotes its exemplar memory, and $\mathcal{P}_l$ is the class distribution of the $l$-th client. $\{\mathcal{P}_l\}_{l=1}^K$ are non-independent and identically distributed (\emph{i.e.}, non-i.i.d.) from each other. At the $t$-th incremental task, the label space $\mathcal{Y}_l^t\subset \mathcal{Y}^t$ of the $l$-th local client is a subset of $\mathcal{Y}^t=\cup_{l=1}^K\mathcal{Y}_l^t$, and it includes $C_l^t$ new classes ($C_l^t \leq C^t$), different from $C_l^p=\sum_{i=1}^{t-1}C_l^i\subset \cup_{j=1}^{t-1}\mathcal{Y}_l^j$ old classes. After loading $\Theta^{r, t}$ and conducting the local training at the $t$-th incremental task, $S_l$ can get a locally updated model $\Theta_{l}^{r, t}$. All locally updated models of selected clients are then uploaded to the global server $S_G$ to be aggregated as the global model $\Theta^{r\!+\!1, t}$ of next round. The global server $\mathcal{S}_G$ then distributes parameters $\Theta^{r\!+\!1, t}$ to local clients for the next global round.

In the FCIL setting, we divide local clients $\{\mathcal{S}_l\}_{l=1}^K$ into three categories (\emph{i.e.}, \{$\mathcal{S}_l\}_{l=1}^K = {\mathbf{S}_o}\cup\mathbf{S}_b\cup \mathbf{S}_n$) in each incremental task. Specifically, $\mathbf{S}_o$ consists of $K_o$ local clients that cannot receive the new data of current task but have the exemplar memory stored via previous learned tasks; $\mathbf{S}_b$ includes $K_b$ clients collecting the new data of current task and the exemplar memory of previous tasks; and $\mathbf{S}_n$ consists of $K_n$ newly-added clients that receive the new data of current task, but have no any exemplar memory of old classes. These clients are dynamically changing as the incremental tasks arriving. Namely, we randomly determine $\{\mathbf{S}_o, \mathbf{S}_b, \mathbf{S}_n\}$ at each global round, and $\mathbf{S}_n$ are irregularly added at any global round in the FCIL. It causes the gradual increase of $K=K_o+K_b+K_n$ in streaming tasks.

Moreover, we have no any prior knowledge about the number of streaming tasks $T$, data distributions $\{\mathcal{P}_l\}_{l=1}^K$, when to collect new classes or add new local clients. The goal of FCIL is to effectively train a global model $\Theta^{R, T}$ to learn new classes consecutively while alleviating the catastrophic forgetting on old classes with the requirement of privacy preservation, via communicating the local model parameters with the global central server $\mathcal{S}_G$.

\begin{figure*}[t]
	\centering
	\includegraphics[width =495pt, height=155pt]
	{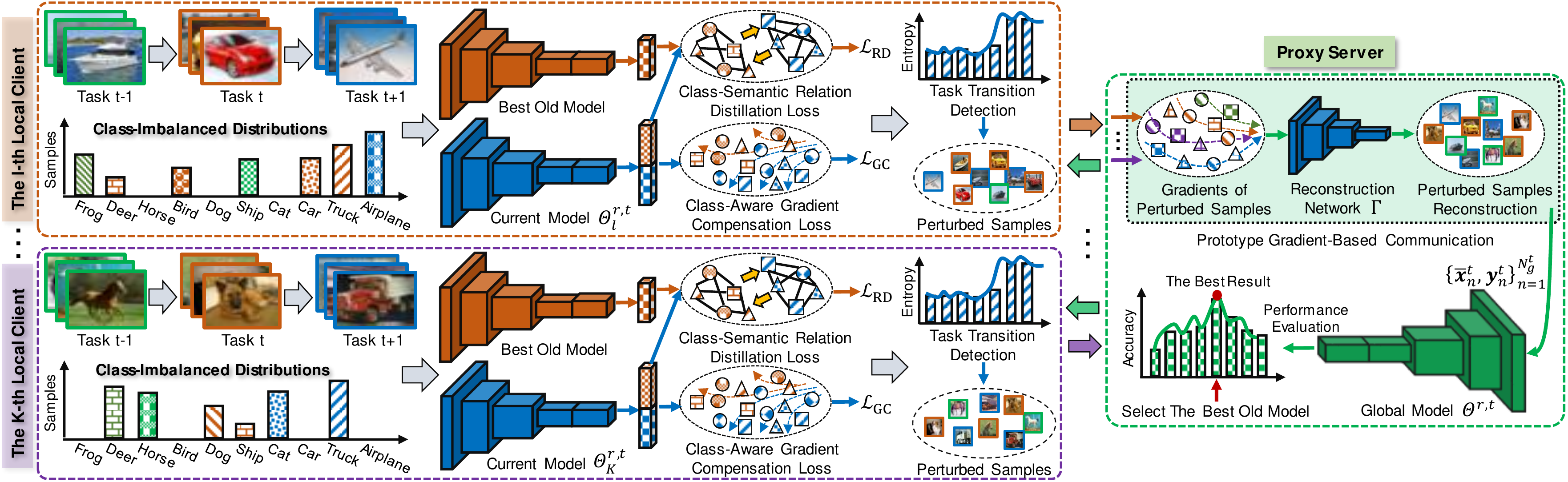}
	\vspace{-20pt}
	\caption{Overview of our GLFC model. It mainly consists of a \textit{class-aware gradient compensation loss} $\mathcal{L}_{\mr{GC}}$ and a \textit{class-semantic relation distillation loss} $\mathcal{L}_{\mr{RD}}$ to overcome local catastrophic forgetting caused by class imbalance at the local side, and a \textit{proxy server} $\mathcal{S}_P$ to address global catastrophic forgetting brought by non-i.i.d. class imbalance across clients, where a prototype gradient-based communication mechanism between $\mathcal{S}_P$ and clients is developed for their private communication while selecting the best old model for $\mathcal{L}_{\mr{RD}}$.}
	\label{fig:overview_of_our_model}
	\vspace{-10pt}
\end{figure*}

\section{The Proposed GLFC Model}
The overview of our model is depicted in Figure~\ref{fig:overview_of_our_model}. To address the FCIL requirements, our model solves local forgetting via a class-aware gradient compensation loss and a class-semantic relation distillation loss (Section \ref{sec:local_forgetting}), while tackling global forgetting via a proxy server to select the best old model for local clients (Section \ref{sec:global_forgetting}).

\subsection{Local Catastrophic Forgetting Compensation} \label{sec:local_forgetting}
At the $t$-th incremental task, given the $l$-th local client $\mathcal{S}_l\in\mathbf{S}_b$ with the training data $\mathcal{T}_l^t$ of new classes and exemplar memory $\mathcal{M}_l$, the classification loss $\mathcal{L}_{\mr{CE}}$ for a mini-batch $\{\mathbf{X}_{lb}^t, \mathbf{Y}_{lb}^t\} = \{\mf{x}_{li}^t, \mf{y}_{li}^t\}_{i=1}^b\subset \mathcal{T}_l^t\cup\mathcal{M}_l$ is:
\vspace{-2pt}
\begin{align}
\begin{split}
\mathcal{L}_{\mr{CE}} =\frac{1}{b}\sum_{i=1}^b \mathcal{D}_{\mr{CE}}(P_l^t(\mathbf{x}_{li}^t, \Theta^{r, t}), \mathbf{y}_{li}^t),
\end{split}
\end{align}
where $b$ is the batch size, and $\Theta^{r, t}$ is the classification model at the $r$-th global round for the $t$-th task, which is transmitted from global server to local clients. $P_l^t(\mathbf{x}_{li}^t, \Theta^{r, t})\in\mathbb{R}^{C^p+C^t}$ denotes the sigmoid probability predicted via $\Theta^{r, t}$, and $\mathcal{D}_{\mr{CE}}(\cdot, \cdot)$ is the binary cross-entropy loss.

As aforementioned, the class imbalance between old and new classes ($\mathcal{T}_l^t$ and $\mathcal{M}_l$) at the local side enforces the local training to suffer from significant performance degradation (\emph{i.e.}, local catastrophic forgetting) on old classes. To prevent local forgetting, as shown in Figure~\ref{fig:overview_of_our_model}, we develop a class-aware gradient compensation loss and a class-semantic relation distillation loss for local clients, which can correct imbalanced gradient propagation and ensure inter-class semantic consistency across incremental tasks.

\textbf{$\bullet$ Class-Aware Gradient Compensation Loss:} 
After $\mathcal{S}_G$ distributes $\Theta^{r, t}$ to local clients, the class-imbalanced distributions at local side cause imbalanced gradient back-propagation of the last output layer in $\Theta^{r, t}$. It forces the update of local model $\Theta_l^{r, t}$ to perform different learning paces within new classes and different forgetting paces within old classes after local training. This phenomenon heavily worsens the local forgetting on old classes, when new streaming data becomes part of old classes continuously.

As a result, we design a class-aware gradient compensation loss $\mathcal{L}_{\mr{{GC}}}$ to respectively normalize the learning paces of new classes and forgetting paces of old classes via re-weighting their gradient propagation. Specifically, inspired by \cite{wang2021addressing, wang2019eavesdrop}, for a single sample $(\mf{x}_{li}^t, \mf{y}_{li}^t)$ (its ground-truth label is $y_{li}^t$, the one-hot vector of $y_{li}^t$ is $\mf{y}_{li}^t$), we obtain a gradient measurement $\mathcal{G}_{li}^t$ with respect to the $y_{li}^t$-th neuron $\mathcal{N}_{y_{li}^t}^t$ of the last output layer in $\Theta_l^{r, t}$: 
\begin{align}
\begin{split}
\mathcal{G}_{li}^t&= \frac{\partial\mathcal{D}_{\mr{CE}}(P_l^t(\mathbf{x}_{li}^t, \Theta_l^{r, t}), \mathbf{y}_{li}^t)}{\partial\mathcal{N}_{y_{li}^t}^t} \\ &= P_{l}^t(\mf{x}_{li}^t, \Theta_l^{r, t})_{y_{li}^t} - 1,
\label{equ:gradient_measurement}
\end{split}
\end{align}
where $P_{l}^t(\mf{x}_{li}^t, \Theta_l^{r, t})_{y_{li}^t}$ is the $y_{li}^t$-th softmax probability of the $i$-th sample. 

To normalize the learning paces of new classes and forgetting paces of old classes, we perform separate gradient normalization for old and new classes, and utilize it to re-weight $\mathcal{L}_{\mr{CE}}$. Given a mini-batch $\{\mathbf{x}_{li}^t,\mathbf{y}_{li}^t\}_{i=1}^b$, we define 
\begin{equation}
\begin{split}
&\mathcal{G}_n = \frac{1}{\sum_{i=1}^b \mathbb{I}_{\mf{y}_{li}^t\in\mathcal{Y}_l^t}} \sum\nolimits_{i=1}^b |\mathcal{G}_{li}^t|\cdot \mathbb{I}_{\mf{y}_{li}^t\in\mathcal{Y}_l^t}, \\ &\mathcal{G}_o = \frac{1}{\sum_{i=1}^b \mathbb{I}_{\mf{y}_{li}^t\in\cup_{j=1}^{t-1}\mathcal{Y}_l^j}} \sum\nolimits_{i=1}^b |\mathcal{G}_{li}^t|\cdot \mathbb{I}_{\mf{y}_{li}^t\in\cup_{j=1}^{t-1}\mathcal{Y}_l^j},
\end{split}
\end{equation}
as the gradient means for new and old classes, where $\mathbb{I}_{(\cdot)}$ is the indicator function that if the subscript condition is true, ${\mathbb{I}}_{(\mr{True})} = 1$; otherwise, ${\mathbb{I}}_{(\mr{False})} = 0$. Thus, the re-weighted $\mathcal{L}_{\mr{CE}}$ loss is formulated as follows: 
\begin{align}	
\mathcal{L}_{\mr{GC}} = \frac{1}{b}\sum_{i=1}^b \frac{|\mathcal{G}_{li}^t|}{\bar{\mathcal{G}}_i} \cdot \mathcal{D}_{\mr{CE}}(P_l^t(\mathbf{x}_{li}^t, \Theta_l^{r, t}), \mathbf{y}_{li}^t),
\label{equ:gradient_compensation}
\end{align}
where  $\bar{\mathcal{G}}_i = \mathbb{I}_{\mf{y}_{li}^t\in\mathcal{Y}_l^t} \cdot \mathcal{G}_n+\mathbb{I}_{\mf{y}_{li}^t\in\cup_{j=1}^{t-1}\mathcal{Y}_l^j} \cdot \mathcal{G}_o$. For instance, when the $i$-th sample $\mf{x}_{li}^t$ belongs to new classes, $\bar{\mathcal{G}}_i$ will be $\mathcal{G}_n$, otherwise $\mathcal{G}_o$.

\textbf{$\bullet$ Class-Semantic Relation Distillation Loss:}
During the training of local model $\Theta_l^{r, t}$ initialized as the current global model $\Theta^{r, t}$, the probability predicted by $\Theta_l^{r, t}$ indicates the inter-class semantic similarity relations. To ensure the inter-class semantic consistency across different incremental tasks, we design a class-semantic relation distillation loss $\mathcal{L}_{\mr{RD}}$ by considering the underlying relations among old and new classes. 
As depicted in Figure~\ref{fig:overview_of_our_model}, we respectively forward a mini-batch dataset $\{\mathbf{X}_{lb}^t, \mathbf{Y}_{lb}^t\}$ into the stored old model $\Theta_l^{t-1}$ and current local model $\Theta_l^{r, t}$, and obtain the corresponding predicted probabilities $P_l^{t-1}(\mathbf{X}_{lb}^t, \Theta_l^{t-1})\in\mathbb{R}^{b\times C^p}$ of old classes and $P_l^t(\mathbf{X}_{lb}^t, \Theta_l^{r, t})\in\mathbb{R}^{b\times (C^p+C^t)}$ of old and new classes. These probabilities reflect the inter-class relations between old and new classes.
Different from existing knowledge distillation strategies \cite{Ahn_2021_ICCV, Hu_2021_CVPR, chen2021learning, deng2021unbiased, DBLP:conf/icml/LiuXL0GS20} that only ensure old classes' semantic consistency among $\Theta_l^{t-1}$ and $\Theta_l^{r, t}$, we consider inter-class relations between old and new classes simultaneously via optimizing $\mathcal{L}_{\mr{RD}}$. Namely, we utilize a variant of one-hot encoding labels $\mathbf{Y}_{lb}^t\in\mathbb{R}^{b\times(C^p+C^t)}$ by replacing the first $C^p$ dimensions of $\mathbf{Y}_{lb}^t$ with $P_l^{t-1}(\mathbf{X}_{lb}^t, \Theta_l^{t-1})$, and denote this variant as $Y_l^t(\mathbf{X}_{lb}^t, \Theta_l^{t-1})\in\mathbb{R}^{b\times(C^p+C^t)}$. Obviously, $Y_l^t(\mathbf{X}_{lb}^t, \Theta_l^{t-1})$ effectively indicates the inter-class semantic similarity relations for both old and new classes via smoothing the one-hot labels. Thus, we formulate $\mathcal{L}_{\mr{RD}}$ as follows: 
\begin{align}	
\mathcal{L}_{\mr{RD}} = \mathcal{D}_{\mr{KL}}(P_l^t(\mathbf{X}_{lb}^t, \Theta_l^{r, t})|| Y_l^t(\mathbf{X}_{lb}^t, \Theta_l^{t-1})),
\label{equ:relation_distillation}
\end{align}
where $\mathcal{D}_{\mr{KL}}(\cdot||\cdot)$ is the Kullback-Leibler divergence. Overall, the optimization objective for the $l$-th local client is:
\begin{align}	
\mathcal{L}_l = \lambda_1\mathcal{L}_{\mr{GC}} + \lambda_2\mathcal{L}_{\mr{RD}},
\label{equ:overall_objective_local_client}
\end{align}
where $\lambda_1, \lambda_2$ are hyper-parameters. We update local model $\Theta_l^{r, t}$ via optimizing Eq.~\eqref{equ:overall_objective_local_client}, and then aggregate all local models in global server $S_G$ to obtain the global model $\Theta^{r\!+\!1, t}$ of the next round. When $t \!=\! 1$, there is no old model $\Theta_l^{t-1}$ to perform $\mathcal{L}_{\mr{RD}}$, and we set $\lambda_1 \!=\! 1.0, \lambda_2 \!=\! 0$, otherwise, $\lambda_1 \!=\! 0.5, \lambda_2 \!=\! 0.5$. Note that local clients in $\mathbf{S}_o$ and $\mathbf{S}_n$ have the same objective (\emph{i.e.}, Eq.~\eqref{equ:overall_objective_local_client}) with the clients in $\mathbf{S}_b$, except for the definition of $\bar{\mathcal{G}}_i$ in Eq.~\eqref{equ:gradient_compensation}. $\bar{\mathcal{G}}_i$ is always set as $\mathcal{G}_o$ for $\mathbf{S}_o$ and $\mathcal{G}_n$ for $\mathbf{S}_n$.

$\bullet$ \textbf{Task Transition Detection:}
When optimizing Eq.~\eqref{equ:overall_objective_local_client}, it is essential for local clients to know when new classes arrive, then update exemplar memory $\mathcal{M}_l$ and store old classification model $\Theta_l^{t-1}$ used for $\mathcal{L}_{\mr{RD}}$. However, in the FCIL, we have no prior knowledge about when local clients receive the data of new classes. To tackle this issue, a trivial solution is to identify whether the labels of training data have been seen before. However, it cannot determine if the newly received labels are from new classes or the old classes observed by other local clients due to non-i.i.d. setting of class distributions. Another intuitive solution is to use performance degradation as the signal of collecting new classes. This solution is infeasible in the FCIL, since the random selection of $\{\mathbf{S}_o, \mathbf{S}_b, \mathbf{S}_n\}$ and their non-i.i.d. class distributions can cause sharp performance degradation, even though without receiving new classes.

To this end, we propose a task transition detection mechanism to accurately identify when local clients receive new classes. Specifically, at the $r$-th global round, every client computes the average entropy $\mathcal{H}_l^{r, t}$ via the received global model $\Theta^{r, t}$ on its current training data $\mathcal{T}_l^t$:
\begin{equation}	
\mathcal{H}_l^{r, t} = \frac{1}{N_l^t}\sum_{i=1}^{N_l^t}\mathcal{I}(P_l^t(\mf{x}_{li}^t, \Theta^{r, t})),
\label{equ:task_transition}
\end{equation}
where $\mathcal{I}(\cdot)=\sum_i p_i\log p_i$ is the entropy function. When $\mathcal{H}_l^{r, t}$ encounters a sudden rise and satisfies $\mathcal{H}_l^{r, t}-\mathcal{H}_l^{r-1, t}\geq r_h$, we argue that the local clients are receiving new classes, and update $t$ by $t\leftarrow t + 1$. Then they can update memory $\mathcal{M}_l$ and store old model $\Theta_l^{t-1}$. We empirically set $r_h=1.2$.

\subsection{Global Catastrophic Forgetting Compensation} \label{sec:global_forgetting}
Although Eq.~\eqref{equ:overall_objective_local_client} could tackle local catastrophic forgetting brought by the class imbalance at the local side, it cannot address heterogeneous forgetting from other local clients (\emph{i.e.}, global catastrophic forgetting). In other words, the non-i.i.d. class-imbalanced distributions across local clients result in certain global catastrophic forgetting on old classes, worsening the local catastrophic forgetting further. 
Thus, it is necessary to address the heterogeneous forgetting across clients from the global perspective. As aforementioned, the proposed class-semantic relation distillation loss $\mathcal{L}_{\mr{RD}}$ in Eq.~\eqref{equ:relation_distillation} requires a stored old classification model $\Theta_l^{t-1}$ of previous tasks to distill inter-class relations. A better $\Theta_l^{t-1}$ can globally increase the distillation gain from previous tasks, strengthening the memory of old classes in a global view.
As a result, the selection of $\Theta_l^{t-1}$ plays an important role in global catastrophic forgetting compensation, which should be considered from a global perspective.

However, in the FCIL, it is difficult to select the best $\Theta_l^{t-1}$ due to the privacy protection. The intuitive solution is that every client stores its best old model $\{\Theta_l^{t-1}\}_{t=2}^T$ for each task during the $(t\!-\!1)$-th task with the training data $\mathcal{T}_l^{t-1}$. Unfortunately, this solution considers the selection of $\Theta_l^{t-1}$ from a local perspective, and cannot guarantee the selected $\Theta_l^{t-1}$ has the best memory for all old classes, since each local client has only a subset of old classes (non-i.i.d.).
To this end, we employ a proxy server $\mathcal{S}_P$ to select the best $\Theta^{t-1}$ for all clients from a global perspective, as depicted in Figure~\ref{fig:overview_of_our_model}. 
Specifically, when local clients have identified new classes (\emph{i.e.}, $\mathcal{T}_l^t$) at the beginning of the $t$-th task via task transition detection, they will transmit perturbed prototype samples of new classes to $\mathcal{S}_P$ via a prototype gradient-based communication mechanism. After receiving these gradients, $\mathcal{S}_P$ reconstructs the perturbed prototype samples, and utilizes them to monitor the performance of global model $\Theta^{r, t}$ (received from $\mathcal{S}_G$) until the best one is found. When stepping at the next task ($t\!+\!1$), $\mathcal{S}_P$ will distribute the best $\Theta^{r, t}$ to local clients, and local clients regard it as the best old model to perform $\mathcal{L}_{\mr{RD}}$.

$\bullet$ \textbf{Prototype Gradient-Based Communication:}
Given the $l$-th local client $\mathcal{S}_l\in\mathbf{S}_b\cup\mathbf{S}_n$ that receives the training data $\mathcal{T}_l^t$ of new classes for the $t$-th task, $\mathcal{S}_l$ identifies new classes via task transition detention. Then $\mathcal{S}_l$ selects only one representative prototype sample $\mf{x}_{lc^*}^t$ from $\mathcal{T}_l^t$ for each new class ($c=C_l^p\!+\!1, \cdots, C_l^p\!+\!C_l^t$), where the feature of $\mf{x}_{lc^*}^t$ is closest to the mean embedding of all samples belonging to the $c$-th class in the latent feature space. We then feed these prototype samples and their labels $\{\mf{x}_{lc^*}^t, \mf{y}_{lc^*}^t\}_{c={C_l^p+1}}^{C_l^p+C_l^t}\subset\mathcal{T}_l^t$ into a $L$-layer gradient encoding network $\Gamma=\{\mathcal{W}_i\}_{i=1}^L$ to compute the gradient $\nabla\Gamma_{lc}$, where $\Gamma$ is much shallower than $\Theta_l^{r, t}$ for communication efficiency, and $\mathcal{W}_i$ is the parameters of the $i$-th layer. The $i$-th element $\nabla_{\mathcal{W}_i}\Gamma_{lc}$ of $\nabla\Gamma_{lc}$ is defined as $\nabla_{\mathcal{W}_i}\Gamma_{lc}= \nabla_{\mathcal{W}_i} \mathcal{D}_{\mr{CE}}(P_l^t(\mf{x}_{lc^*}^t, \Gamma), \mf{y}_{lc^*}^t)$, where $P_l^t(\mf{x}_{lc^*}^t, \Gamma)$ is the probability predicted via $\Gamma$. Then $\mathcal{S}_l$ transmit $C_l^t$ gradients $\{\nabla\Gamma_{lc}\}_{c={C_l^p+1}}^{C_l^p+C_l^t}$ to $\mathcal{S}_P$ for the reconstruction of prototype samples.

\begin{table*}[t]
	\centering
	\setlength{\tabcolsep}{2.8mm}
	\caption{Performance comparisons between our model and other baseline methods on CIFAR-100 \cite{krizhevsky2009learning} with 10 incremental tasks. }
	\vspace{-10pt}
	\scalebox{0.885}{
		\begin{tabular}{|c|cccccccccc|>{\columncolor{lightgray}}c|>{\columncolor{lightgray}}c|}
			\hline
			Methods & 10 & 20 & 30 & 40 & 50 & 60 & 70 & 80 & 90 & 100 & Avg. & $~\triangle~$ \\
			\hline
			iCaRL \cite{Rebuffi_2017_CVPR} + FL & 89.0 & 55.0 & 57.0 & 52.3 & 50.3 & 49.3 & 46.3 & 41.7 & 40.3 & 36.7 & 51.8 & $\Uparrow$15.1 \\ 
			BiC \cite{wu2019large} + FL & 88.7 & 63.3 & 61.3 & 56.7 & 53.0 & 51.7 & 48.0 & 44.0 & 42.7 & 40.7 & 55.0 & $\Uparrow$11.9 \\ 
			PODNet \cite{10.1007/978-3-030-58565-5_6} + FL & 89.0 & 71.3 & 69.0 & 63.3 & 59.0 & 55.3 & 50.7 & 48.7 & 45.3 & 45.0 & 59.7 & $\Uparrow$7.2 \\
			DDE \cite{Hu_2021_CVPR} + iCaRL \cite{Rebuffi_2017_CVPR} + FL & 88.0 & 70.0 & 67.3 & 62.0 & 57.3 & 54.7 & 50.3 & 48.3 & 45.7 & 44.3 & 58.8 & $\Uparrow$8.1 \\
			GeoDL \cite{Christian2021MGeoCont} + iCaRL \cite{Rebuffi_2017_CVPR} + FL & 87.0 & 76.0 & 70.3 & 64.3 & 60.7 & 57.3 & 54.7 & 50.3 & 48.3 & 46.3 & 61.5 & $\Uparrow$5.4 \\
			SS-IL \cite{Ahn_2021_ICCV} + FL & 88.3 & 66.3 & 54.0 & 54.0 & 44.7 & 54.7 & 50.0 & 47.7 & 45.3 & 44.0 & 54.9 & $\Uparrow$12.0 \\
			\hline
			\hline
			Ours-w/oCGC & 89.0 & 80.0 & 75.0 & 70.0 & 63.3 & 62.0 & 57.0 & 54.7 & 50.3 & 46.0 & 60.1 & $\Uparrow$6.8 \\
			Ours-w/oCRD & 89.0 & 80.3 & 76.0 & 71.0 & 64.0 & 65.0 & 57.7 & 56.0 & 51.0 & 48.3 & 65.8 & $\Uparrow$1.1 \\
			Ours-w/oPRS & 88.0 & 80.3 & 75.0 & 70.3 & 62.0 & 63.0 & 58.0 & 54.3 & 49.0 & 45.7 & 64.6 & $\Uparrow$2.3 \\
			Ours & 90.0 & 82.3 & 77.0 & 72.3 & 65.0 & 66.3 & 59.7 & 56.3 & 50.3 & 50.0 & \textbf{66.9} & -- \\ 
			
			\hline
		\end{tabular}
	} 	
	\label{tab:exp_CIFAR_100}
	\vspace{-10pt}
\end{table*}

$\mathcal{S}_P$ randomly shuffles all received gradients from selected clients of this global round to construct a gradient pool $\nabla\Gamma^t = \cup_l\{\nabla\Gamma_{lc}\}_{c={C_l^p+1}}^{C_l^p+C_l^t}$, and we assume there are $N_g^t$ gradients in this pool. This shuffling operation prevents $\mathcal{S}_P$ from tracking certain selected clients by remarking special gradient distributions. For the $n$-th element $\nabla\Gamma_n^t$ of $\nabla\Gamma^t$, we can obtain its corresponding ground-truth label $y_n^t$ (with one-hot encoding label $\mf{y}_n^t$) via observing the gradient symbol of the last layer in $\nabla\Gamma_n^t$ (proposed in \cite{zhao2020idlg, wang2021addressing}). Given a dummy sample $\bar{\mf{x}}_n^t$ initialized by a standard Gaussian $\mathcal{N}(0, 1)$, we forward all pairs of $\{\bar{\mf{x}}_n^t, \nabla\Gamma_n^t, \mf{y}_n^t\}_{n=1}^{N_g^t}$ into $\Gamma=\{\mathcal{W}_i\}_{i=1}^L$ that is same as the gradient encoding network used by local clients, to recover prototype samples for each new class. The reconstruction loss $\mathcal{L}_{\mr{RT}}$ and the update of $\bar{\mf{x}}_t^n$ are expressed as follows:
\begin{align}
\!\mathcal{L}_{\mr{RT}} &= \sum_{i=1}^{L} \|\nabla_{\mathcal{W}_i} \mathcal{D}_{\mr{CE}}(P^t(\bar{\mf{x}}_n^t, \Gamma), \mf{y}_n^t) - \nabla_{\mathcal{W}_i}\Gamma_n^t \|^2, \\ 
\bar{\mf{x}}_n^t &\leftarrow \bar{\mf{x}}_n^t - \eta\nabla_{\bar{\mf{x}}_n^t}\mathcal{L}_{\mr{RT}},
\label{equ:reconstruction_loss}
\end{align}
where $P^t(\bar{\mf{x}}_n^t, \Gamma)$ is the probability predicted via $\Gamma$. $\eta$ denotes the learning rate to update $\bar{\mf{x}}_n^t$.

$\bullet$ \textbf{Selection of The Best Old Model:}
$\mathcal{S}_P$ can only receive gradients from local clients at the first round of the $t$-th task, when they detect new classes. Then $\mathcal{S}_P$ reconstructs $N_g^t$ prototype samples of new classes and their labels (\emph{i.e.}, $\{\bar{\mf{x}}_n^t, \mf{y}_n^t\}_{n=1}^{N_g^t}$) via optimizing Eq.~\eqref{equ:reconstruction_loss}. At the $t$-th task, $\mathcal{S}_P$ forwards these reconstructed samples into the global model $\Theta^{r, t}$ (received from $\mathcal{S}_G$) to select the best $\Theta^{t}$ via evaluating which model has the best accuracy, until receiving gradients of new classes from the next task. At every global round starting from the second task, $\mathcal{S}_P$ distributes the best models of the last task and current task (\emph{i.e.}, $\Theta^{t-1}$ and $\Theta^{t}$), to all selected clients. If these selected clients detect new classes from $\mathcal{T}_l^{t+1}$ at the $t$-th task, they will set $\Theta^{t}$ as the old model $\Theta_l^{t-1}$, otherwise, $\Theta^{t-1}$ is set as $\Theta_l^{t-1}$ to perform $\mathcal{L}_{\mr{RD}}$.

$\bullet$ \textbf{Perturbed Prototype Samples Construction:}
Although the network $\Gamma$ is only privately accessible to $\mathcal{S}_P$ and local clients, malicious attackers may steal $\Gamma$ and these gradients to reconstruct raw prototype sample $\{\mf{x}_{lc^*}^t, \mf{y}_{lc^*}^t\}\in\mathcal{T}_l^t$ of the $l$-th local client. To achieve privacy preservation, we propose to add perturbations to these prototype samples. The attackers can get little useful information from the perturbed prototype samples even if they can reconstruct them.

\begin{figure}[t]
	\centering
	\includegraphics[width=0.47\textwidth]
	{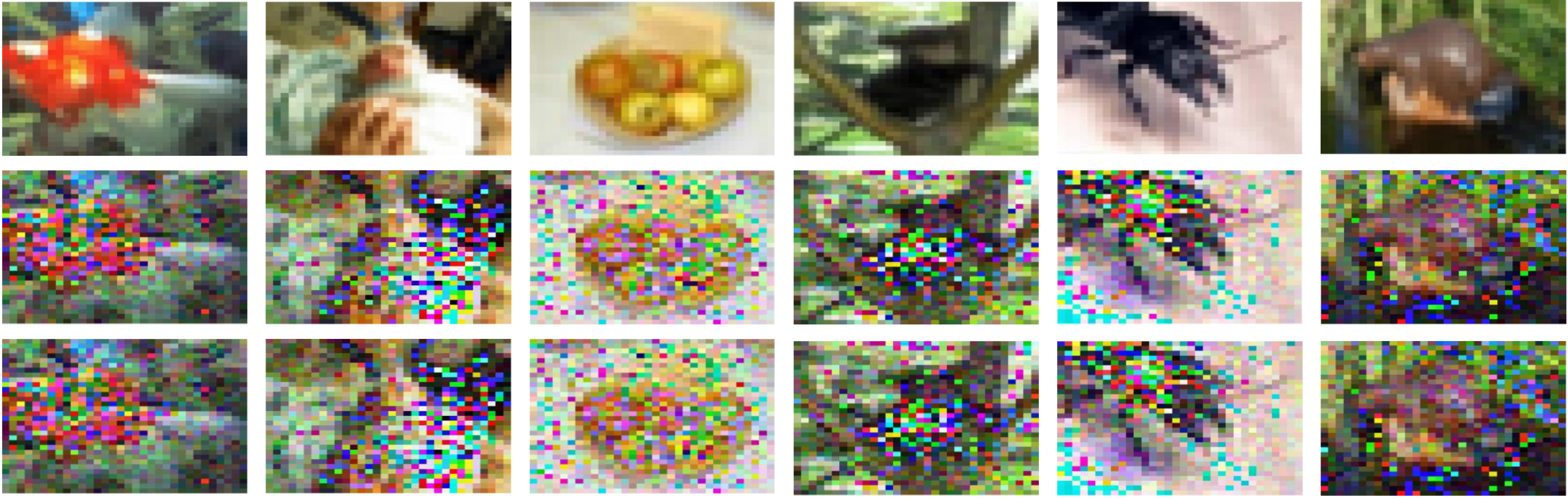}
	\vspace{-8pt}
	\caption{Visualization of original prototype samples (top row), perturbed prototype samples (middle row), and reconstructed prototype samples via proxy server (bottom row) in CIFAR-100 \cite{krizhevsky2009learning}.  }
	\label{fig:prototype_samples}
	\vspace{-10pt}
\end{figure}

To be specific, given a prototype sample $\{\mf{x}_{lc^*}^t, \mf{y}_{lc^*}^t\}\in\mathcal{T}_l^t$, we forward it into the local model $\Theta_l^{r, t}$ that has been trained via Eq.~\eqref{equ:overall_objective_local_client}, and apply back-propagation to update this sample. In order to produce perturbed prototype sample, we introduce a Gaussian noise into the latent feature of prototype sample, and then update $\mf{x}_{lc^*}^t$ via Eq.~\eqref{equ:perturbation_reconstruction}:
\begin{align}
\mathcal{L}_{\mr{GP}} &= \mathcal{D}_{\mr{CE}}(P_l^t(\Phi(\mf{x}_{lc^*}^t) + \gamma \mathcal{N}(0, \sigma^2), \Theta_l^{r, t}), \mf{y}_{lc^*}^t), \\
\mf{x}_{lc^*}^t &\leftarrow \mf{x}_{lc^*}^t - \eta \nabla_{\mf{x}_{lc^*}^t}\mathcal{L}_{\mr{GP}},
\label{equ:perturbation_reconstruction}
\end{align}
where $\Phi(\mf{x}_{lc^*}^t)$ denotes the latent feature of $\mf{x}_{lc^*}^t$, and $P_l^t(\Phi(\mf{x}_{lc^*}^t) + \gamma \mathcal{N}(0, \sigma^2), \Theta_l^{r, t})$ is the probability predicted via $\Theta_l^{r, t}$ when adding Gaussian noise $\mathcal{N}(0, \sigma^2)$ to $\Phi(\mf{x}_{lc^*}^t)$. $\sigma^2$ represents the variance of features of all samples belonging to $\mf{y}_{lc^*}^t$, and we empirically set $\gamma=0.1$ to control the effect of Gaussian noise in this paper. Some reconstructed prototype samples are visualized in Figure~\ref{fig:prototype_samples}.

\subsection{Optimization Pipeline of Our GLFC Model}
Starting from the first incremental task, all clients are required to compute the average entropy of their private training data via Eq.~\eqref{equ:task_transition} at the beginning of each global round, and follow iCaRL \cite{Rebuffi_2017_CVPR} to update their exemplar memory $\mathcal{M}_l$. For each global training round, the central server $\mathcal{S}_G$ randomly selects a set of local clients to conduct local training. After that, when the selected clients identify new classes via the task transition detection strategy, they will construct perturbed prototype samples of these new classes and share the corresponding gradients to the proxy server $\mathcal{S}_P$ via the prototype gradient-based communication mechanism. After receiving these gradients, $\mathcal{S}_P$ reconstructs these prototype samples, and utilizes them to select the best global model $\Theta^t$ until collecting gradients next time. Starting from the second task ($t\!=\!2$), $\mathcal{S}_P$ will distribute best models of the last and current task (\emph{i.e.}, $\Theta^{t-1}$, and $\Theta^t$) to selected clients. 
Then the $l$-th client uses $\Theta^{t-1}$ as its $\Theta_l^{t-1}$ to update the current local model $\Theta_l^{r, t}$ via optimizing Eq.~\eqref{equ:overall_objective_local_client}, when it doesn't detect new classes via task transition detection. Otherwise, it uses $\Theta^{t}$ to train the current local model $\Theta_l^{r, t}$. Finally, $\mathcal{S}_G$ aggregates the updated local models $\Theta_l^{r, t}$ to get the global model $\Theta^{r+1, t}$ of next ground. 
The optimization pipeline is provided in supplementary material. 

\begin{table*}[t]
	\centering
	\setlength{\tabcolsep}{2.8mm}
	\caption{Performance comparisons between our model and other baseline methods on ImageNet-Subset \cite{5206848} with 10 incremental tasks. }
	\vspace{-10pt}
	\scalebox{0.885}{
		\begin{tabular}{|c|cccccccccc|>{\columncolor{lightgray}}c|>{\columncolor{lightgray}}c|}
			\hline
			Methods & 10 & 20 & 30 & 40 & 50 & 60 & 70 & 80 & 90 & 100 & Avg. & $~\triangle~$ \\
			\hline
			iCaRL \cite{Rebuffi_2017_CVPR} + FL & 74.0 & 62.3 & 56.3 & 47.7 & 46.0 & 40.3 & 37.7 & 34.3 & 33.3 & 32.7 & 46.5 & $\Uparrow$10.5 \\ 
			BiC \cite{wu2019large} + FL & 74.3 & 63.0 & 57.7 & 51.3 & 48.3 & 46.0 & 42.7 & 37.7 & 35.3 & 34.0 & 49.0 & $\Uparrow$8.0 \\ 
			PODNet \cite{10.1007/978-3-030-58565-5_6} + FL & 74.3 & 64.0 & 59.0 & 56.7 & 52.7 & 50.3 & 47.0 & 43.3 & 40.0 & 38.3 & 52.6 & $\Uparrow$4.4 \\ 
			DDE \cite{Hu_2021_CVPR} + iCaRL \cite{Rebuffi_2017_CVPR} + FL & 76.0 & 57.7 & 58.0 & 56.3 & 53.3 & 50.7 & 47.3 & 44.0 & 40.7 & 39.0 & 52.3 & $\Uparrow$4.7 \\
			GeoDL \cite{Christian2021MGeoCont} + iCaRL \cite{Rebuffi_2017_CVPR} + FL & 74.0 & 63.3 & 54.7 & 53.3 & 50.7 & 46.7 & 41.3 & 39.7 & 38.3 & 37.0 & 50.0 & $\Uparrow$7.0 \\
			SS-IL \cite{Ahn_2021_ICCV} + FL & 69.7 & 60.0 & 50.3 & 45.7 & 41.7 & 44.3 & 39.0 & 38.3 & 38.0 & 37.3 & 46.4 & $\Uparrow$10.6 \\
			\hline
			\hline
			Ours-w/oCGC & 74.0 & 67.0 & 61.0 & 60.0 & 57.0 & 53.7 & 50.0 & 47.0 & 42.0 & 39.3 & 55.1 & $\Uparrow$1.9\\
			Ours-w/oCRD & 76.0 & 56.0 & 53.7 & 45.0 & 46.0 & 43.0 & 42.0 & 39.3 & 37.0 & 35.3 & 47.3 & $\Uparrow$9.7 \\
			Ours-w/oPRS & 73.0 & 64.0 & 62.3 & 57.3 & 54.0 & 50.3 & 46.7 & 43.0 & 40.0 & 37.3 & 52.8 & $\Uparrow$4.2\\
			Ours & 73.0 & 69.3 & 68.0 & 61.0 & 58.3 & 54.0 & 51.3 & 48.0 & 44.3 & 42.7 & \textbf{57.0} & -- \\ 
			
			\hline
		\end{tabular}
	} 	
	\label{tab:exp_ImageNet_Subset}
	\vspace{-10pt}
\end{table*}
\begin{table*}[t]
	\centering
	\setlength{\tabcolsep}{2.8mm}
	\caption{Comparisons of the first 10 tasks on TinyImageNet \cite{Tiny_imagenet} with 20 tasks, where the rest comparisons are in supplementary material. }
	\vspace{-10pt}
	\scalebox{0.885}{
		\begin{tabular}{|c|cccccccccc|>{\columncolor{lightgray}}c|>{\columncolor{lightgray}}c|}
			\hline
			Methods & 10 & 20 & 30 & 40 & 50 & 60 & 70 & 80 & 90 & 100 & Avg. & $~\triangle~$ \\
			\hline
			iCaRL \cite{Rebuffi_2017_CVPR} + FL & 67.0 & 59.3 & 54.0 & 48.3 & 46.7 & 44.7 & 43.3 & 39.0 & 37.3 & 33.0 & 47.3 & $\Uparrow$7.6 \\ 
			BiC \cite{wu2019large} + FL & 67.3 & 59.7 & 54.7 & 50.0 & 48.3 & 45.3 & 43.0 & 40.7 & 38.0 & 33.7 & 48.1 & $\Uparrow$6.8 \\ 
			PODNet \cite{10.1007/978-3-030-58565-5_6} + FL & 69.0 & 59.3 & 55.0 & 51.7 & 50.0 & 46.7 & 43.7 & 41.0 & 39.3 & 38.0 & 49.4 & $\Uparrow$5.5 \\ 
			DDE \cite{Hu_2021_CVPR} + iCaRL \cite{Rebuffi_2017_CVPR} + FL & 70.0 & 59.3 & 53.3& 51.0 & 48.3 & 45.7 & 42.3 & 40.0 & 38.0 & 36.3 & 48.4 & $\Uparrow$6.5 \\
			GeoDL \cite{Christian2021MGeoCont} + iCaRL \cite{Rebuffi_2017_CVPR} + FL & 66.3 & 56.7 & 51.0 & 49.7 & 44.7 & 42.3 & 41.0 & 39.0 & 37.3 & 35.0 & 46.3 & $\Uparrow$8.6 \\
			SS-IL \cite{Ahn_2021_ICCV} + FL & 66.7 & 54.0 & 47.7 & 45.3 & 42.3 & 42.0 & 40.7 & 38.0 & 36.0 & 34.3 & 44.7 & $\Uparrow$10.2  \\
			\hline
			\hline
			Ours-w/oCGC & 67.7 & 60.3 & 57.7 & 55.0 & 51.0 & 49.0 & 48.0 & 45.7 & 44.3 & 42.0 & 52.1 & $\Uparrow$2.8\\
			Ours-w/oCRD & 68.0 & 60.0 & 53.0 & 47.3 & 42.0 & 39.0 & 37.3 & 35.3 & 33.7 & 32.0 & 44.8 & $\Uparrow$10.1 \\
			Ours-w/oPRS & 67.3 & 59.7 & 55.0 & 51.3 & 50.7 & 48.0 & 46.3 & 43.3 & 41.7 & 40.3 & 50.3 & $\Uparrow$4.6\\
			Ours & 68.7 & 63.3 & 61.7 & 57.3 & 56.0 & 53.0 & 50.3 & 47.7 & 46.3 & 45.0 & \textbf{54.9} & -- \\
			
			\hline
		\end{tabular}
	} 	
	\label{tab:exp_TinyImageNet}
	\vspace{-15pt}
\end{table*}

\section{Experiments}\label{sec:experiments}

\subsection{Implementation Details}
\vspace{-5pt}
We use three datasets: CIFAR-100 \cite{krizhevsky2009learning, DBLP:conf/icml/FangLLL021}, ImageNet-Subset \cite{5206848}, and TinyImageNet \cite{Tiny_imagenet} in our experiments. For a fair comparison with baseline class-incremental learning methods \cite{Rebuffi_2017_CVPR, wu2019large, 10.1007/978-3-030-58565-5_6, Hu_2021_CVPR, Christian2021MGeoCont, Ahn_2021_ICCV} in the FCIL setting,
we follow the same protocols proposed by \cite{Rebuffi_2017_CVPR, wu2019large} to set incremental tasks, utilize the identical class order generated from iCaRL \cite{Rebuffi_2017_CVPR}, and employ the same backbone (\emph{i.e.}, ResNet-18 \cite{7780459}) as the classification model \cite{chen2019two}. The SGD optimizer whose learning rate is 2.0 is used to train all models. The exemplar memory $\mathcal{M}_l$ of each client is set as 2,000 for all streaming tasks. A shallow LeNet \cite{726791} with only 4 layers is used as the network $\Gamma$. We employ a SGD optimizer with the learning rate as 0.1 to construct perturbed samples for local clients, while utilizing a L-BFGS optimizer with the learning rate as 1.0 to reconstruct prototype samples for proxy server. We initialize the number of local clients as 30 in the first incremental task, introduce 10 additional new local clients as the learning tasks arrive consecutively. At each global round, we randomly select 10 clients to conduct 20-epoch local training. Each client randomly receives 60\% classes from the label space of its seen task. We run our experiments for 3 times with 3 random seeds (2021, 2022, 2023)
and report the averaged results. 
Please refer to more details in supplementary material. 

\subsection{Performance Comparison}
This section shows comparison experiments to illustrate the effectiveness of our GLFC model, as shown in Tables~\ref{tab:exp_CIFAR_100}, \ref{tab:exp_ImageNet_Subset}, \ref{tab:exp_TinyImageNet}, where `$\triangle$' denotes the improvements of our model compared with other comparison methods. We observe that our model outperforms existing class-incremental methods \cite{Rebuffi_2017_CVPR, wu2019large, 10.1007/978-3-030-58565-5_6, Hu_2021_CVPR, Christian2021MGeoCont, Ahn_2021_ICCV} in the FCIL setting by a margin of 4.4\%$\sim$15.1\% in terms of average accuracy. It validates that our model could enable local clients to collaboratively train a global class-incremental model. Moreover, our model has stable performance improvement in comparison with other methods for all incremental tasks, which verifies the effectiveness of our model to address the forgetting in FCIL.

\begin{figure*}[h]
	\centering
	\includegraphics[width =495pt, height=145pt]
	{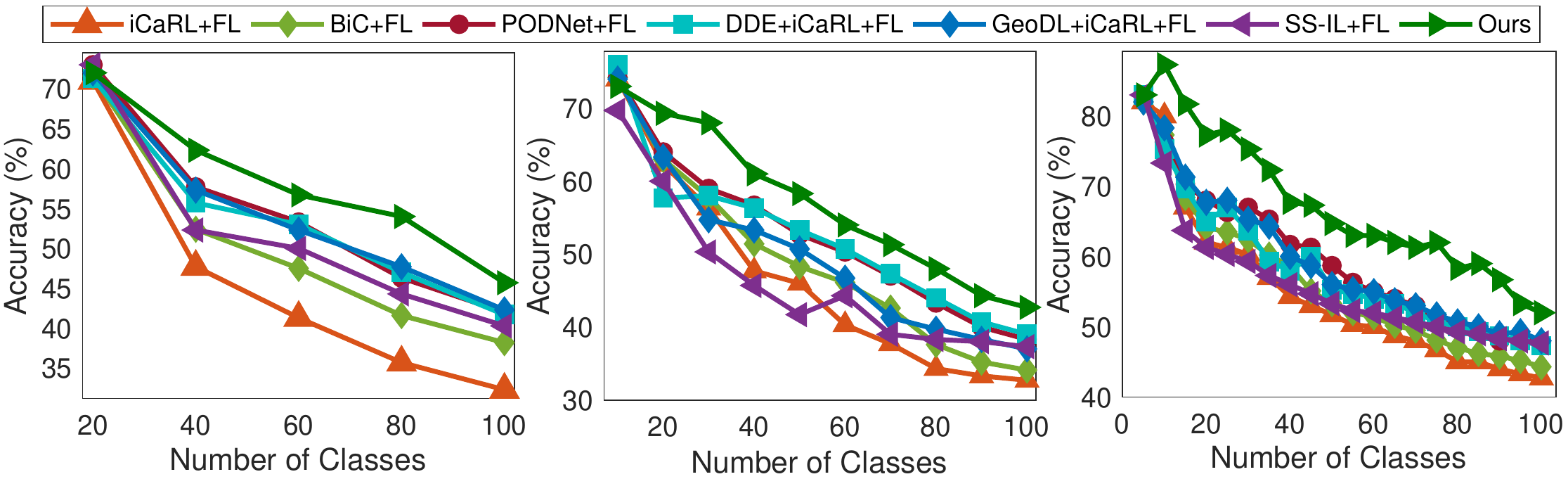}
	\vspace{-20pt}
	\caption{Qualitative analysis of different incremental tasks on CIFAR-100 \cite{krizhevsky2009learning} when $T=5$ (left), $T=10$ (middle) and $T=20$ (right). }
	\label{fig:incremental_tasks_CIFAR_100}
	\vspace{-10pt}
\end{figure*}
\begin{figure*}[t]
	\centering
	\includegraphics[width =495pt, height=145pt]
	{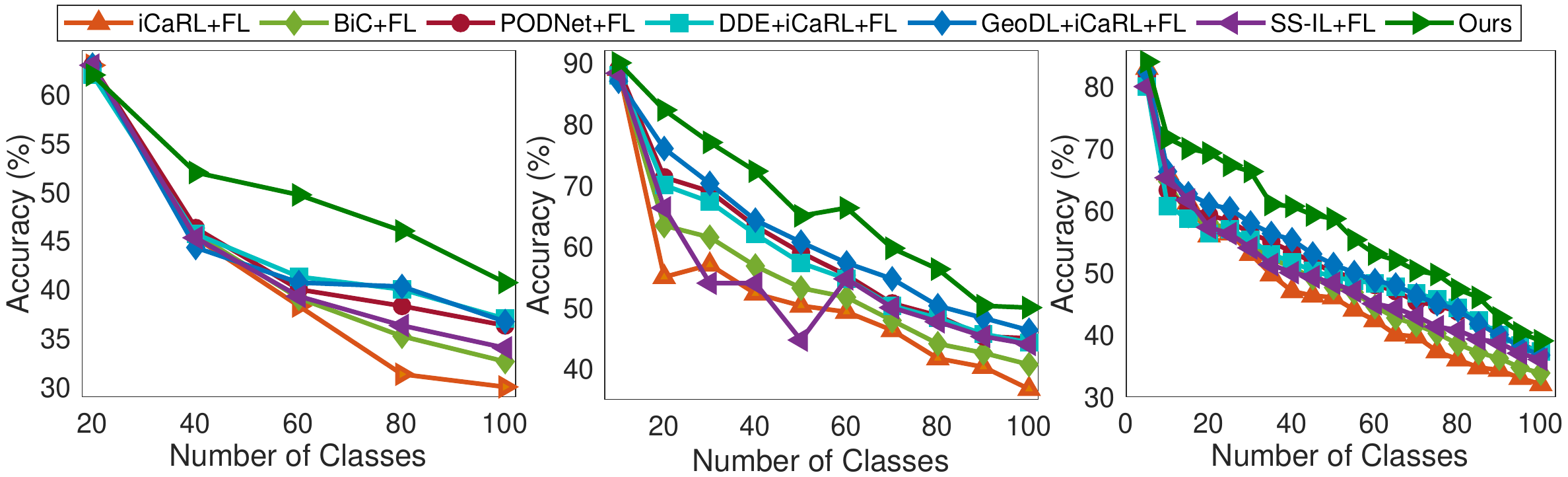}
	\vspace{-20pt}
	\caption{Qualitative analysis of incremental tasks on ImageNet-Subset \cite{5206848} when $T=5$ (left), $T=10$ (middle) and $T=20$ (right). }
	\label{fig:incremental_tasks_SubImageNet}
	\vspace{-15pt}
\end{figure*}

\subsection{Ablation Studies}
\vspace{-5pt}
As shown in Tables~\ref{tab:exp_CIFAR_100}, \ref{tab:exp_ImageNet_Subset}, \ref{tab:exp_TinyImageNet}, we investigate the effects of each module in our model via ablation studies. Ours-w/oCGC, Ours-w/oCRD and Ours-w/oPRS denote the performance of our model without using $\mathcal{L}_{\mr{GC}}$, $\mathcal{L}_{\mr{RD}}$ and the proxy server $\mathcal{S}_P$, where Ours-w/oCGC and Ours-w/oCRD utilize $\mathcal{L}_{\mr{CE}}$ and the knowledge distillation proposed in iCaRL \cite{Rebuffi_2017_CVPR} for a replacement. Compared with Ours, the performance of Ours-w/oCGC, Ours-w/oCRD and Ours-w/oPRS degrades evidently with a range of 1.1\%$\sim$10.1\%. It verifies the effectiveness of all modules to cooperate together. The ablation performance verifies all modules are essential to train a global class-incremental model. The proxy server is also essential to select the best old model via evaluating reconstructed samples (shown in Figure~\ref{fig:prototype_samples}).

\begin{table}[t]
	\centering
	\setlength{\tabcolsep}{2.85mm}
	\caption{Qualitative analysis of different exemplar memories in local clients on CIFAR-100 \cite{krizhevsky2009learning} when $T=5$. }
	\vspace{-10pt}
	\scalebox{0.97}{
		\begin{tabular}{|c|ccccc|>{\columncolor{lightgray}}c|>{\columncolor{lightgray}}c|}
			\hline
			$\mathcal{M}_l$ & 20 & 40 & 60 & 80 & 100 & Avg.\\
			\hline
			500 & 71.3 & 52.7 & 47.0 & 45.0 & 38.3 & 50.9 \\
			1000 & 71.0 & 56.3 & 52.7 & 50.0 & 42.3 & 54.5 \\
			1500 & 71.0 & 58.0 & 54.3 & 53.0 & 44.0 & 56.1 \\
			2000 & 72.0 & 62.3 & 56.7 & 54.0 & 45.7 & 58.1 \\
			\hline
		\end{tabular} 
	} 	
	\label{tab:analysis_different_memories_cifar}
	\vspace{-10pt}
\end{table}

\subsection{Qualitative Analysis of Incremental Tasks}
In this section, we conduct qualitative analysis of various incremental tasks ($T\!=\!5,10,20$) on benchmark datasets to validate the superior performance of GLFC, as shown in Figures~\ref{fig:incremental_tasks_CIFAR_100}, \ref{fig:incremental_tasks_SubImageNet}. According to these curves, we can easily observe that our model performs better than other competing baseline methods \cite{Rebuffi_2017_CVPR, wu2019large, 10.1007/978-3-030-58565-5_6, Hu_2021_CVPR, Christian2021MGeoCont, Ahn_2021_ICCV} for all incremental tasks, in the settings with different number of tasks ($T\!=\!5,10,20$). It demonstrates that the GLFC model enables multiple local clients to learn new classes in a streaming manner while addressing local and global forgetting.

\subsection{Qualitative Analysis of Exemplar Memory}
As shown in Table~\ref{tab:analysis_different_memories_cifar}, we study the effects of different exemplar memories on the performance of our GLFC model, by respectively setting $\mathcal{M}_l$ as $\{500, 1000, 1500, 2000\}$ on CIFAR-100 \cite{krizhevsky2009learning}. From the results in Table~\ref{tab:analysis_different_memories_cifar}, we can conclude that the better performance of GLFC model on learning new classes in a streaming manner will be, the larger size of exemplar memory $\mathcal{M}_l$ is. It validates that storing more training data of old classes at the local side could strengthen the memory ability of our proposed GLFC model for old classes. Moreover, the presented results also illustrate the effectiveness of our GLFC model about identifying new classes via the task transition detection mechanism and updating the exemplar memory.

\section{Conclusion}
In this paper, we propose a practical Federated Class-Incremental Learning (FCIL) problem, and develop a novel Global-Local Forgetting Compensation (GLFC) model to address the local and global catastrophic forgetting in FCIL. Specifically, a class-aware gradient compensation loss and a class-semantic relation distillation loss are designed to locally address the catastrophic forgetting, by correcting the imbalanced gradient propagation and ensuring consistent inter-class relations across tasks. We also employ a proxy server to tackle global forgetting by selecting the best old model for preserving the memory of old classes. Extensive experiments on representative benchmark datasets demonstrate the effectiveness of our proposed GLFC model.

\section*{Acknowledgement}
This work was partially supported by National Nature Science Foundation of China under Grant 62003336; National Science Foundation of US under Grants 1834701, 2016240; and research awards from Facebook, Google, PlatON Network, and General Motors. 

{\small
\bibliographystyle{ieee_fullname}
\bibliography{CVPR2022_FCIL}
}

\clearpage
\appendix

\section{Implementation Details}
\label{sm_implementation_datails}

\subsection{Datasets}
\textbf{CIFAR-100} \cite{krizhevsky2009learning} is composed of 60,000 color images from 100 classes. Each class has 500 training and 100 evaluation samples with the size $32\times 32$. 

\textbf{ImageNet-Subset} \cite{5206848} is a subset of ImageNet \cite{5206848}, and it includes 100 classes sampled in the same way as \cite{Hu_2021_CVPR}. We split each class into 500 training and 100 test samples with the size $224\times 224$. 

\textbf{TinyImageNet} \cite{Tiny_imagenet} includes 100,000 samples for 200 classes, and each sample is downsized to $64\times 64$. Each class has 500 training and 50 test samples.

\subsection{Experimental Settings}
For the federated learning setting, in the first task, there are total 30 local clients, and for each global round, we randomly select 10 clients to conduct 20-epoch local training. After the local training, these clients will share their updated models to participate in the global aggregation of this round. When the number of streaming tasks is $T = 10$, for CIFAR-100 and ImageNet-Subset, each task includes 10 new classes for 10 global rounds, and each task transition will introduce 10 additional new clients. For TinyImageNet, each task includes 20 new classes for the same 10 global rounds, and each task transition also includes 10 new clients. Therefore, in the last task of $T=10$, the total number of local clients is 120, and we also randomly select 10 clients to perform global aggregation. For the case of $T=20$, each task has 5 classes with 10 global rounds of training for CIFAR-100 and ImageNet-Subset, and the number of classes will be 10 for TinyImageNet. Note that the number of newly introduced clients is 5 now at each task transition. We also conduct experiments of $T=5$, CIFAR-100 and ImageNet-Subset contain 20 classes for each task, while TinyImageNet has 40 classes per task. For all three datasets, each task covers 20 global rounds and there will be 20 new clients joining in the framework at each task transition. For building the non-i.i.d. setting, every client can only own 60\% classes of the label space in the current task, and these classes are randomly selected. During the task transition global round, we assume $90\%$ existing clients are from $\mathcal{S}_b$, while the resting $10\%$ clients are from $\mathcal{S}_o$.

For fair comparisons with other class-incremental learning methods in the FCIL setting, we follow the same protocols proposed by \cite{Rebuffi_2017_CVPR, wu2019large} to split classes into incremental tasks, utilize the identical class order generated from iCaRL \cite{Rebuffi_2017_CVPR}. Moreover, we use ResNet-18 as the classification model. As for the gradient encoding network, we use a shallow LeNet with only 4 layers. We use a SGD optimizer whose initial learning rate is 2.0 to train all classification models, and the learning rate is divided by 5, 25 and 125 when the accumulated local epochs of a task hit 100, 150 and 180, respectively. What's more, the optimizer of back-propagation perturbation generation is SGD with learning rate as 0.1, and the sample reconstruction optimization happened on the proxy server utilizes L-BFGS with learning rate as 1.0 for saving memory storage. The batch size is 128, and the exemplar memory $\mathcal{M}_l$ of each client has the sample size of 2,000 during all streaming tasks. As for the optimization iterations, the prototype perturbation generation has 100 iterations, and prototype sample reconstruction conducts 200 iterations for each gradient. We repeatedly run our experiments for three times with three random seeds (2021, 2022, 2023) and report the average results in our comparison experiments.

\begin{algorithm}[!t]
\small
\caption{Optimization Pipeline of Our Model.}
\label{alg:pipline_GLFC}
\LinesNumbered 
\textbf{Given:} At the $r$-th global round and the $t$-th task (assume $t \geq  2$), central server $\mathcal{S}_G$ randomly selects a set of clients $\{\mathcal{S}_{s_1}, \mathcal{S}_{s_2}, ..., \mathcal{S}_{s_m}\}$ with size as $m$;
The selected clients have their local training data $\{\mathcal{T}_{s_1}^t, \mathcal{T}_{s_2}^t, \cdots, \mathcal{T}_{s_m}^t\}$ and local exemplar memories $\{\mathcal{M}_{s_1}, \mathcal{M}_{s_2}, \cdots, \mathcal{M}_{s_m}\}$; $\mathcal{S}_G$ sends the latest global classification model $\Theta^{r, t}$ to all selected clients; The gradient encoding model $\Gamma$ and the proxy server $\mathcal{S}_P$;

\textcolor{blue}{\textbf{All Clients:}} \\
\For{$\mathcal{S}_l$ in $\{\mathcal{S}_1, \mathcal{S}_2, \cdots, \mathcal{S}_K\}$}{
    Use $\mathcal{T}_{l}^t$ to compute average entropy $\mathcal{H}_l^{r, t}$ via Eq.~\eqref{equ:task_transition}; \\
    Apply iCaRL~\cite{Rebuffi_2017_CVPR} on $\mathcal{T}_{l}^t$ to update $\mathcal{M}_l$; \\
}

\textcolor{blue}{\textbf{Selected Clients:}} \\
Receive $\Theta^{r, t}$ from $\mathcal{S}_G$ as local classification model;\\
Receive $\Theta^{t-1}, \Theta^{t}$ from $\mathcal{S}_P$\;
\For{$\mathcal{S}_{l}$ in $\{\mathcal{S}_{s_1}, \mathcal{S}_{s_2}, \cdots, \mathcal{S}_{s_m}\}$}{
    \emph{Task} = False\;
    \If{$\mathcal{H}_{l}^{r, t} - \mathcal{H}_{l}^{r-1, t} \geq r_h$}{
        \emph{Task} = True\;
    }
    \If{Task = \emph{True}}{
        $\Theta_{l}^{t-1} = \Theta^{t}$\;
    }
    \Else{
        $\Theta_{l}^{t-1} = \Theta^{t-1}$\;
    }
    \For{$\{\mathbf{X}_{lb}^t, \mathbf{Y}_{lb}^t\}$ in $\mathcal{T}_{l}^t \cup \mathcal{M}_{l}$}{
        Update local model $\Theta_{l}^{r, t}$ via optimizing Eq.~\eqref{equ:overall_objective_local_client}; \\
    }
    \If{Task = \emph{True}}{
        $\nabla \Gamma_{l}^t = \{\}$\;
        \For{$c$ in $[C_{l}^p+1, C_{l}^p+C_{l}^t]$}{
            Generate perturbed sample $(\mathbf{x}_{lc^*}^t, \mathbf{y}_{lc^*}^t)$\;
            Compute gradient $\nabla \Gamma_{lc} = \cup_{\mathcal{W}_i} \nabla_{\mathcal{W}_i} \mathcal{D}_{\mr{CE}}(P_{l}^t(\mathbf{x}_{lc^*}^t, \Gamma), \mathbf{y}_{lc^*}^t)$\;
            $\nabla \Gamma_{l}^t \leftarrow \nabla \Gamma_{l}^t \cup \nabla \Gamma_{lc}$
        }
        Send $\nabla \Gamma_{l}^t$ to the proxy server $\mathcal{S}_P$\;
    }
}
\textcolor{blue}{\textbf{Proxy Server:}} \\
Receive $\nabla \Gamma^t = \{\nabla \Gamma_{s_1}^t, \nabla \Gamma_{s_2}^t, ..., \nabla \Gamma_{s_m}^t\}$ from selected local clients, and there are $N_g^t$ gradients in $\nabla \Gamma^t$\;
Receive $\Theta^{r, t}$ from $\mathcal{S}_G$ as local classification model;\\
\If{$N_g^t \neq 0$}{
    Shuffle the gradient pool $\nabla \Gamma^t$\;
    $\{\bar{\mathbf{X}}_P^t, \mathbf{Y}_P^t\} = \{\}$\;
    \For{$n = 1, \cdots, N_g^t$}{
        Reconstruct $\bar{\mathbf{x}}_n^t$ via optimizing Eq.~\eqref{equ:reconstruction_loss};
        $\{\bar{\mathbf{X}}_P^t, \mathbf{Y}_P^t\} \leftarrow \{\bar{\mathbf{X}}_P^t, \mathbf{Y}_P^t\} \cup (\bar{\mathbf{x}}_n^t, \mathbf{y}_n^t)$\;
    }
}
Forward $\{\bar{\mathbf{X}}_P^t, \mathbf{Y}_P^t\}$ to $\Theta^{r, t}$ and get the best $\Theta^t$\;
Distribute $\Theta^{t-1}$ and $\Theta^t$ to all selected local clients\;
\end{algorithm}

\subsection{Comparison Methods} 
This paper is the first exploration to address the federated class-incremental learning (FCIL) problem, and there is not any baseline method that is built on similar settings. Therefore, for fair comparisons, we compare our GLFC model with several state-of-the-art class-incremental methods (\emph{i.e.}, iCaRL \cite{Rebuffi_2017_CVPR}, BiC \cite{wu2019large}, PODNet \cite{10.1007/978-3-030-58565-5_6}, DDE \cite{Hu_2021_CVPR}, GeoDL \cite{Christian2021MGeoCont} and SS-IL \cite{Ahn_2021_ICCV} under the federated learning (FL) settings, to validate the effectiveness of our proposed GLFC model. Besides, top-1 accuracy metric is employed to evaluate the performance of other comparison methods and our proposed GLFC model.

\section{Optimization Pipeline of Our GLFC Model}
\label{sm_pipeline}
Starting from the first incremental task, all clients are required to compute the average entropy of their private training data via Eq.~\eqref{equ:task_transition} at the beginning of each global round, and follow iCaRL \cite{Rebuffi_2017_CVPR} to update their exemplar memory $\mathcal{M}_l$. For each global training round, the central server $\mathcal{S}_G$ randomly selects a set of local clients to conduct local training. After that, when the selected clients identify new classes via the task transition detection strategy, they will construct perturbed prototype samples of these new classes and share the corresponding gradients to the proxy server $\mathcal{S}_P$ via the prototype gradient-based communication mechanism. After receiving these gradients, $\mathcal{S}_P$ reconstructs these prototype samples, and utilizes them to select the best global model $\Theta^t$ until collecting gradients next time. Starting from the second task ($t\!=\!2$), $\mathcal{S}_P$ will distribute best models of the last and current task (\emph{i.e.}, $\Theta^{t-1}$, and $\Theta^t$) to selected clients. 
Then the $l$-th client uses $\Theta^{t-1}$ as its $\Theta_l^{t-1}$ to update the current local model $\Theta_l^{r, t}$ via optimizing Eq.~\eqref{equ:overall_objective_local_client}, when it doesn't detect new classes via task transition detection. Otherwise, it can use $\Theta^{t}$ to train the current local model $\Theta_l^{r, t}$. Finally, $\mathcal{S}_G$ aggregates the updated local models $\Theta_l^{r, t}$ to get the global model $\Theta^{r+1, t}$ of next ground. 
The detailed optimization pipeline is provided in Algorithm~\ref{alg:pipline_GLFC}.

\begin{table*}[!]
	\centering
	\setlength{\tabcolsep}{2.8mm}
	\caption{Comparisons of the first 10 tasks between our model and other baseline methods on TinyImageNet \cite{Tiny_imagenet} with 20 incremental tasks. }
	\vspace{-10pt}
	\scalebox{0.885}{
		\begin{tabular}{|c|cccccccccc|>{\columncolor{lightgray}}c|>{\columncolor{lightgray}}c|}
			\hline
			Methods & 10 & 20 & 30 & 40 & 50 & 60 & 70 & 80 & 90 & 100 & Avg. & $~\triangle~$ \\
			\hline
			iCaRL \cite{Rebuffi_2017_CVPR} + FL & 67.0 & 59.3 & 54.0 & 48.3 & 46.7 & 44.7 & 43.3 & 39.0 & 37.3 & 33.0 & 47.3 & $\Uparrow$7.6 \\ 
			BiC \cite{wu2019large} + FL & 67.3 & 59.7 & 54.7 & 50.0 & 48.3 & 45.3 & 43.0 & 40.7 & 38.0 & 33.7 & 48.1 & $\Uparrow$6.8 \\ 
			PODNet \cite{10.1007/978-3-030-58565-5_6} + FL & 69.0 & 59.3 & 55.0 & 51.7 & 50.0 & 46.7 & 43.7 & 41.0 & 39.3 & 38.0 & 49.4 & $\Uparrow$5.5 \\ 
			DDE \cite{Hu_2021_CVPR} + iCaRL \cite{Rebuffi_2017_CVPR} + FL & 70.0 & 59.3 & 53.3& 51.0 & 48.3 & 45.7 & 42.3 & 40.0 & 38.0 & 36.3 & 48.4 & $\Uparrow$6.5 \\
			GeoDL \cite{Christian2021MGeoCont} + iCaRL \cite{Rebuffi_2017_CVPR} + FL & 66.3 & 56.7 & 51.0 & 49.7 & 44.7 & 42.3 & 41.0 & 39.0 & 37.3 & 35.0 & 46.3 & $\Uparrow$8.6 \\
			SS-IL \cite{Ahn_2021_ICCV} + FL & 66.7 & 54.0 & 47.7 & 45.3 & 42.3 & 42.0 & 40.7 & 38.0 & 36.0 & 34.3 & 44.7 & $\Uparrow$10.2  \\
			\hline
			\hline
			Ours-w/oCGC & 67.7 & 60.3 & 57.7 & 55.0 & 51.0 & 49.0 & 48.0 & 45.7 & 44.3 & 42.0 & 52.1 & $\Uparrow$2.8\\
			Ours-w/oCRD & 68.0 & 60.0 & 53.0 & 47.3 & 42.0 & 39.0 & 37.3 & 35.3 & 33.7 & 32.0 & 44.8 & $\Uparrow$10.1 \\
			Ours-w/oPRS & 67.3 & 59.7 & 55.0 & 51.3 & 50.7 & 48.0 & 46.3 & 43.3 & 41.7 & 40.3 & 50.3 & $\Uparrow$4.6\\
			Ours & 68.7 & 63.3 & 61.7 & 57.3 & 56.0 & 53.0 & 50.3 & 47.7 & 46.3 & 45.0 & \textbf{54.9} & -- \\
			
			\hline
		\end{tabular}
	} 	
	\label{tab:exp_TinyImageNet_first10}
\end{table*}
\begin{table*}[!]
	\centering
	\setlength{\tabcolsep}{2.8mm}
	\caption{Comparisons of the last 10 tasks  between our model and other baseline methods on TinyImageNet \cite{Tiny_imagenet} with 20 incremental tasks. }
	\vspace{-10pt}
	\scalebox{0.885}{
		\begin{tabular}{|c|cccccccccc|>{\columncolor{lightgray}}c|>{\columncolor{lightgray}}c|}
			\hline
			Methods & 110 & 120 & 130 & 140 & 150 & 160 & 170 & 180 & 190 & 200 & Avg. & $~\triangle~$ \\
			\hline
			iCaRL \cite{Rebuffi_2017_CVPR} + FL & 32.0 & 30.3 & 28.0 & 27.0 & 26.3 & 25.3 & 24.7 & 24.0 & 22.7 & 22.0 & 26.2 & $\Uparrow$11.0 \\ 
			BiC \cite{wu2019large} + FL & 32.7 & 32.3 & 30.3 & 29.0 & 27.7 & 27.3 & 26.0 & 25.7 & 24.3 & 23.3 & 27.9 & $\Uparrow$9.3 \\ 
			PODNet \cite{10.1007/978-3-030-58565-5_6} + FL & 37.0 & 35.7 & 34.7 & 34.0 & 33.0 & 32.3 & 31.0 & 30.0 & 29.3 & 28.0 & 32.5 & $\Uparrow$4.7 \\ 
			DDE \cite{Hu_2021_CVPR} + iCaRL \cite{Rebuffi_2017_CVPR} + FL & 35.0 & 33.7 & 32.0 & 31.0 & 30.3 & 30.0 & 28.7 & 28.3 & 27.3 & 26.0 & 30.2 & $\Uparrow$7.0 \\ 
			GeoDL \cite{Christian2021MGeoCont} + iCaRL \cite{Rebuffi_2017_CVPR} + FL & 33.7 & 32.0 & 31.0 & 30.3 & 28.7 & 28.0 & 27.3 & 26.3 & 25.0 & 24.7 & 28.7 & $\Uparrow$8.5 \\ 
			SS-IL \cite{Ahn_2021_ICCV} + FL & 33.0 & 31.0 & 29.3 & 28.3 & 27.7 & 27.0 & 26.3 & 26.0 & 25.0 & 24.3 & 27.8 & $\Uparrow$9.4 \\ 
			\hline
			\hline
			Ours-w/oCGC & 40.7 & 38.3 & 37.3 & 36.0 & 35.3 & 33.7 & 33.0 & 31.7 & 30.3 & 29.0 & 34.5 & $\Uparrow$2.7 \\ 
			Ours-w/oCRD & 30.7 & 29.7 & 29.3 & 28.0 & 27.7 & 27.0 & 25.7 & 25.0 & 24.0 & 22.7 & 27.0 & $\Uparrow$10.2 \\
			Ours-w/oPRS & 39.0 & 38.0 & 37.3 & 36.3 & 34.7 & 33.3 & 31.7 & 31.0 & 30.3 & 28.7 & 34.0 & $\Uparrow$3.2 \\ 
			Ours & 42.7 & 41.0 & 40.0 & 39.3 & 38.0 & 36.7 & 35.3 & 34.0 & 33.0 & 31.7 & \textbf{37.2} & -- \\ 
			
			\hline
		\end{tabular}
	} 	
	\label{tab:exp_TinyImageNet_last10}
\end{table*}
\begin{table*}[!]
	\centering
	\setlength{\tabcolsep}{2.8mm}
	\caption{Performance comparisons between our model and other baseline methods on TinyImageNet \cite{Tiny_imagenet} with 10 incremental tasks.}
	\vspace{-10pt}
	\scalebox{0.885}{
		\begin{tabular}{|c|cccccccccc|>{\columncolor{lightgray}}c|>{\columncolor{lightgray}}c|}
			\hline
			Methods & 20 & 40 & 60 & 80 & 100 & 120 & 140 & 160 & 180 & 200 & Avg. & $~\triangle~$ \\
			\hline
			iCaRL \cite{Rebuffi_2017_CVPR} + FL & 63.0 & 53.0 & 48.0 & 41.7 & 38.0 & 36.0 & 33.3 & 30.7 & 29.7 & 28.0 & 40.1 & $\Uparrow$7.8 \\ 
			BiC \cite{wu2019large} + FL & 65.3 & 52.7 & 49.3 & 46.0 & 40.3 & 38.3 & 35.7 & 33.0 & 31.7 & 29.0 & 42.1 & $\Uparrow$5.8 \\ 
			PODNet \cite{10.1007/978-3-030-58565-5_6} + FL & 66.7 & 53.3 & 50.0 & 47.3 & 43.7 & 42.7 & 40.0 & 37.3 & 33.7 & 31.3 & 44.6 & $\Uparrow$3.3 \\ 
			DDE \cite{Hu_2021_CVPR} + iCaRL \cite{Rebuffi_2017_CVPR} + FL & 69.0 & 52.0 & 50.7 & 47.0 & 43.3 & 42.0 & 39.3 & 37.0 & 33.0 & 31.3 & 44.5 & $\Uparrow$3.4 \\ 
			GeoDL \cite{Christian2021MGeoCont} + iCaRL \cite{Rebuffi_2017_CVPR} + FL & 66.3 & 54.3 & 52.0 & 48.7 & 45.0 & 42.0 & 39.3 & 36.0 & 32.7 & 30.0 & 44.6 & $\Uparrow$3.3 \\ 
			SS-IL \cite{Ahn_2021_ICCV} + FL & 62.0 & 48.7 & 40.0 & 38.0 & 37.0 & 35.0 & 32.3 & 30.3 & 28.7 & 27.0 & 37.9 & $\Uparrow$10.0 \\ 
			Ours & 66.0 & 58.3 & 55.3 & 51.0 & 47.7 & 45.3 & 43.0 & 40.0 & 37.3 & 35.0 & \textbf{47.9} & -- \\ 
			
			\hline
		\end{tabular}
	} 	
	\label{tab:exp_TinyImageNet_task10}
\end{table*}

\section{Experiments on TinyImageNet Dataset}
\label{sm_additional_experiments}
\subsection{Performance Comparison}
As shown in Tables~\ref{tab:exp_TinyImageNet_first10}, \ref{tab:exp_TinyImageNet_last10}, we present comparison experiments between our model and other baseline class-incremental learning methods on TinyImageNet dataset. The presented results show that our GLFC model significantly outperforms other state-of-the-art comparison methods by 4.7\%$\sim$11.0\% in terms of average accuracy. It illustrates the effectiveness of our model to address both local and global catastrophic forgetting in the FCIL setting. Moreover, the performance of our model is the best among all incremental tasks, and there is a large performance improvement for each incremental task. This phenomenon validates that the proposed proxy server is effective to address global catastrophic forgetting brought by non-i.i.d. class imbalance across clients via prototype sample construction mechanism. Meanwhile, the proposed class-aware gradient compensation
loss and class-semantic relation distillation loss guarantee that our model could effectively alleviate local catastrophic forgetting at local client side.

\subsection{Ablation Studies}
This subsection investigates the effectiveness of different variants of our model on TinyImageNet dataset, as presented in Tables~\ref{tab:exp_TinyImageNet_first10}, \ref{tab:exp_TinyImageNet_last10}. When compared with Ours, Ours-w/oCGC degrades the performance of 2.7\%$\sim$2.8\% in terms of average accuracy, which validates the effectiveness of the class-aware gradient compensation loss to compensate imbalanced gradient propagation. We observe that Ours performs better than Ours-w/oCRD by 10.1\%$\sim$10.2\% in terms of average accuracy. The class-semantic relation distillation loss ensures inter-class semantic consistency across different incremental tasks to address local catastrophic forgetting. Moreover, the performance of Ours-w/oPRS is worse than Ours by 3.2\%$\sim$4.6\% in terms of average accuracy. This performance degradation verifies that global catastrophic forgetting brought by non-i.i.d. class imbalance across clients could be effectively addressed via the proxy server. All proposed modules in our GLFC model could cooperate well to address the FCIL problem. When any one of proposed components is removed, as shown in Tables~\ref{tab:exp_TinyImageNet_first10}, \ref{tab:exp_TinyImageNet_last10}, Ours-w/oCGC, Ours-w/oCRD and Ours-w/oPRS achieve significant performance degradation.

\subsection{Effects of Incremental Tasks}
As presented in Tables~\ref{tab:exp_TinyImageNet_task10}, \ref{tab:exp_TinyImageNet_task5}, in this subsection, we introduce the qualitative analysis of various incremental tasks ($T=5, 10$) on TinyImageNet dataset to validate the effectiveness of the proposed GLFC model. From the results in Tables~\ref{tab:exp_TinyImageNet_task10}, \ref{tab:exp_TinyImageNet_task5}, we observe that the performance of our proposed model has a large improvement (3.2\%$\sim$10.0\% in terms of average accuracy) over other state-of-the-art comparison methods for all incremental tasks. Even though there are different settings with different number of tasks ($T=5, 10$), our proposed GLFC model still has the best performance, which verifies that our model could effectively tackle both local and global catastrophic forgetting in the FCIL setting. Moreover, the significant performance improvement illustrates that our model enables multiple local clients to learn new classes consecutively, while addressing catastrophic forgetting on old learned classes under the privacy preservation and limited memory of local clients. 

\begin{table*}[t]
	\centering
	\setlength{\tabcolsep}{4.8mm}
	\caption{Performance comparisons between our model and other baseline methods on TinyImageNet \cite{Tiny_imagenet} with 5 incremental tasks.}
	\vspace{-10pt}
	\scalebox{0.999}{
		\begin{tabular}{|c|ccccc|>{\columncolor{lightgray}}c|>{\columncolor{lightgray}}c|}
			\hline
			Methods & 40 & 80 & 120 & 160 & 200 & Avg. & $~\triangle~$ \\
			\hline
			iCaRL \cite{Rebuffi_2017_CVPR} + FL & 65.0 & 48.0 & 42.7 & 38.7 & 35.0 & 45.9 & $\Uparrow$5.2  \\
			BiC \cite{wu2019large} + FL & 65.7 & 48.7 & 43.0 & 40.3 & 35.7 & 46.7 & $\Uparrow$4.4  \\ 
			PODNet \cite{10.1007/978-3-030-58565-5_6} + FL & 66.0 & 50.3 & 44.7 & 41.3 & 37.0 & 47.9 & $\Uparrow$3.2  \\
			DDE \cite{Hu_2021_CVPR} + iCaRL \cite{Rebuffi_2017_CVPR} + FL & 63.0 & 51.3 & 45.3 & 41.0 & 36.0 & 47.3 & $\Uparrow$3.8  \\
			GeoDL \cite{Christian2021MGeoCont} + iCaRL \cite{Rebuffi_2017_CVPR} + FL & 65.3 & 50.0 & 45.0 & 40.7 & 36.0 & 47.4 & $\Uparrow$3.7  \\
			SS-IL \cite{Ahn_2021_ICCV} + FL & 65.0 & 42.3 & 38.3 & 35.0 & 30.3 & 42.2 & $\Uparrow$8.9  \\
			Ours & 66.0 & 55.3 & 49.0 & 45.0 & 40.3 & \textbf{51.1} & --  \\

			\hline
		\end{tabular}
	} 	
	\label{tab:exp_TinyImageNet_task5}
\end{table*}

\begin{table*}[t]
	\centering
	\setlength{\tabcolsep}{4.05mm}
	\caption{Qualitative analysis of different exemplar memories in local clients on CIFAR-100 \cite{krizhevsky2009learning} when $T=10$. }
	\vspace{-10pt}
	\scalebox{0.999}{
		\begin{tabular}{|c|cccccccccc|>{\columncolor{lightgray}}c|}
			\hline
			$\mathcal{M}_l$ & 10 & 20 & 30 & 40 & 50 & 60 & 70 & 80 & 90 & 100 & Avg.  \\
			\hline
			500 & 90.0 & 74.3 & 66.0 & 58.3 & 52.0 & 51.0 & 43.0 & 42.0 & 40.0 & 39.3 & 55.6 \\ 
			1000 & 89.0 & 78.3 & 72.0 & 64.3 & 59.7 & 59.0 & 52.3 & 49.3 & 48.7 & 47.7 & 62.0 \\ 
			1500 & 89.3 & 82.0 & 76.0 & 70.0 & 64.0 & 64.0 & 56.3 & 52.7 & 49.3 & 48.7 & 65.2  \\ 
			2000 & 90.0 & 82.3 & 77.0 & 72.3 & 65.0 & 66.3 & 59.7 & 56.3 & 50.3 & 50.0 & \textbf{66.9} \\ 
			
			\hline
		\end{tabular}
	} 	
	\label{tab:analysis_different_memories}
\end{table*}

\section{Qualitative Analysis of Exemplar Memory}
In this subsection, as shown in Table~\ref{tab:analysis_different_memories}, we further conduct extensive experiments ($T=10$) on CIFAR-100 dataset to investigate the effects of different exemplar memories on the performance of our proposed GLFC model when setting $\mathcal{M}_l$ as $\{500, 1000, 1500, 2000\}$. From the presented results in Table~\ref{tab:analysis_different_memories}, we easily observe that our model achieves the better performance for all incremental tasks, when local clients have large memory storage to store the exemplar samples of old classes. Moreover, storing more training data of old classes at the local side could promote the memory replay on old classes, which further addresses catastrophic forgetting at local clients' side for old classes. Besides, it validates that our proposed model is efficient to distinguish new classes via the task transition detection strategy and update the corresponding exemplar memory $\mathcal{M}_l$ at local side. The updated exemplar memory plays an essential role in tackling local catastrophic forgetting on old classes.

\section{Limitation and Societal Impact}
\label{sm_limitation}
This section discusses the limitation for our proposed model and the potential societal impact of this paper.

\subsection{Limitation}
This paper mainly focuses on addressing Federated Class-Incremental Learning (FCIL) problem from the algorithm perspective. In the future, it is necessary to develop mathematical theoretical supports for understanding the FCIL problem and the proposed GLFC model. A possible way to develop mathematical theories for our model is considering existing mathematical explanations of federated learning (FL) and class-incremental learning (CIL) simultaneously. However, 
as we know, there is rare theoretical analysis about CIL. Therefore, it might be tough to propose a brand-new theoretical support to analyze regular CIL problem. Instead, we will try to establish a theoretical analysis for the FCIL problem from a FL perspective in the future work.

\subsection{Potential Societal Impact}
The FCIL problem discussed in our paper doesn't have any negative societal impact. On the contrary, we believe our work can solve real-world problems and bring about extensive benefits. In comparison with standard federated learning (FL), the proposed Federated Class-Incremental Learning (FCIL) is more practical as we assume the data of new classes as well as new clients will indiscriminately and continuously participate in FCIL. To solve the FCIL problem, our proposed GLFC model can enable a global class-incremental learning model to be trained on decentralized devices without data sharing (uploading decentralized data to a central server or data exchange between participated devices). Compared to regular class-incremental learning methods that always need access to the training data, FCIL can protect the private information of participants by remaining the local data where it is collected.

We have faith that the proposed GLFC model can bring beneficial gains to a number of information-sensitive scenarios, such as medical diagnosis, smartphone applications, pharmaceutical companies, and high-technology enterprises, etc. In summary, this work is the first attempt to learn a global class-incremental model in the setting of FL, which expedites the development of FL-based applications with the requirement of privacy preservation.

\end{document}